\documentclass[11pt]{article}

\usepackage[final]{acl}

\usepackage{times}
\usepackage{latexsym}
\usepackage{amsmath}
\usepackage{amssymb}

\usepackage[T1]{fontenc}

\usepackage[utf8]{inputenc}

\usepackage{microtype}

\usepackage{inconsolata}

\usepackage{graphicx}

\title{Efficient Extractive Summarization with MAMBA-Transformer Hybrids for Low-Resource Scenarios}

\author{Nisrine Ait Khayi\\
\small Independent Researcher\\
\small University of Memphis}

\begin{document}
\maketitle
\begin{abstract}
Extractive summarization of long documents is bottlenecked by quadratic complexity, often forcing truncation and limiting deployment in resource-constrained settings. We introduce the first Mamba–Transformer hybrid for extractive summarization, combining the semantic strength of pre-trained transformers with the linear-time processing of state space models. Leveraging Mamba’s ability to process full documents without truncation, our approach preserves context while maintaining strong summarization quality.
The architecture includes: (1) a transformer encoder for sentence-level semantics, (2) a Mamba state space model to capture inter-sentence dependencies efficiently, and (3) a linear classifier for sentence relevance prediction.
Across news, argumentative, and scientific domains under low-resource conditions, our method achieves: (1) large gains over BERTSUM and MATCHSUM, including +0.23 ROUGE-1 on ArXiv and statistically significant improvements on all datasets ($p < 0.001$);(2) consistent advantages across domains, strongest on the longest documents; (3) robust performance with limited training data.
We introduce the first hybrid Transformer–state space architectures for summarization, showing significant ROUGE improvements and efficiency gains in low-resource scenarios; and (4) 24–27\% faster on news summarization (CNN/DailyMail).
\end{abstract}

\section{Introduction}

Text summarization is a crucial task in Natural Language Processing (NLP) that generates concise text containing the most significant information from original documents. News summarization has become critical for information consumption in the digital age, with approaches ranging from extractive methods like \cite{dlljzw25,leluu24}to abstractive techniques \cite{bzz25,zzlflklxzrmz24}.Recent deep neural approaches achieve strong performance but require substantial computational resources and large-scale training data \cite{zlz23,azwzhq22,sl21}. 
However, these advances face critical challenges, particularly in resource-constrained environments. Specifically, transformer-based summarization models rely heavily on self-attention mechanisms and suffer from quadratic computational complexity with respect to sequence length, forcing truncation. Thus, requiring massive computational resources for both training and inference, as well as very large, annotated datasets that are expensive to create.
To address these computational barriers while maintaining effective semantic understanding, we propose, for the first time, a Mamba-Transformer hybrid architecture for extractive summarization. In this approach, pre-trained transformer encoders generate rich sentence-level representations. Then processed through Mamba's linear-complexity state space model to capture inter-sentence dependencies while processing full-length documents without truncation.  This design preserves complete contextual information, reduces inference cost, and proves especially effective for very long documents such as scientific papers, where transformers, often fail to retain all context.
The main contributions of this work are:
\begin{enumerate}
	\item We propose the first Mamba-Transformer hybrid model for extractive summarization.
	\item We demonstrate linear complexity scaling and no truncation, enabling efficient summarization of long documents.
	\item We provide cross-domain low-resource evaluation. Comprehensive experiments on news, argumentative, and scientific datasets .
	\item Showed up to +0.23 ROUGE-1 improvement on Arxiv, significant gains (p < 0.001) and 24-27 \% faster inference on news.
    \item We demonstrate the robustness in long documents settings. Largest gains observed on the longest documents, demonstrating the architecture’s advantage in scenarios where sequence length and context retention are critical.
\end{enumerate}

\section{Related Work}

\subsection{Computational Challenges in Transformer-Based NLP}
Transformer-based models achieve impressive NLP performance but suffer from quadratic computational complexity, limiting deployment in resource-constrained environments \cite{vb21}. Recent efficiency methods include pruning \cite{gda20}, knowledge distillation \cite{zlxlmwot22}, and quantization \cite{jdhllwx24}. We employ quantization-aware fine-tuning for resource-constrained experimentation.

\subsection{State Space Models as Efficient Alternatives}
Mamba State Space models \cite{gd23} offer linear O(L) complexity, making them attractive for sequence modeling. Recent works demonstrate their potential in text ranking \cite{xygs25} and machine translation \cite{pitorroetal24}, matching transformer performance with fewer resources. However, their effectiveness for extractive summarization remains underexplored.

\subsection{Extractive Summarization}
Transformer-based approaches dominate extractive summarization. BERTSUM \cite{l19} established sentence-level extraction paradigms, while MATCHSUM \cite{zlcwqh20}introduces semantic matching. These methods are not competitive in low-resource settings. Recent methods like COLO \cite{azwzhq22} and DiffuSum \cite{zlz23} achieve strong performance but inherit quadratic complexity, limiting low-resource applicability. Recent works address low-resource scenarios through disentangled representation learning \cite{yswcl23} and paraphrasing reformulation \cite{twwcgq23}. Our contribution differs by addressing computational efficiency at the architectural level, enabling the first hybrid approach that maintains competitive performance while achieving linear complexity.
\section{Method}
Figure 1 depicts the architecture of our proposed method.

\begin{figure}[t]
    \centering
    \includegraphics[width=0.9\linewidth]{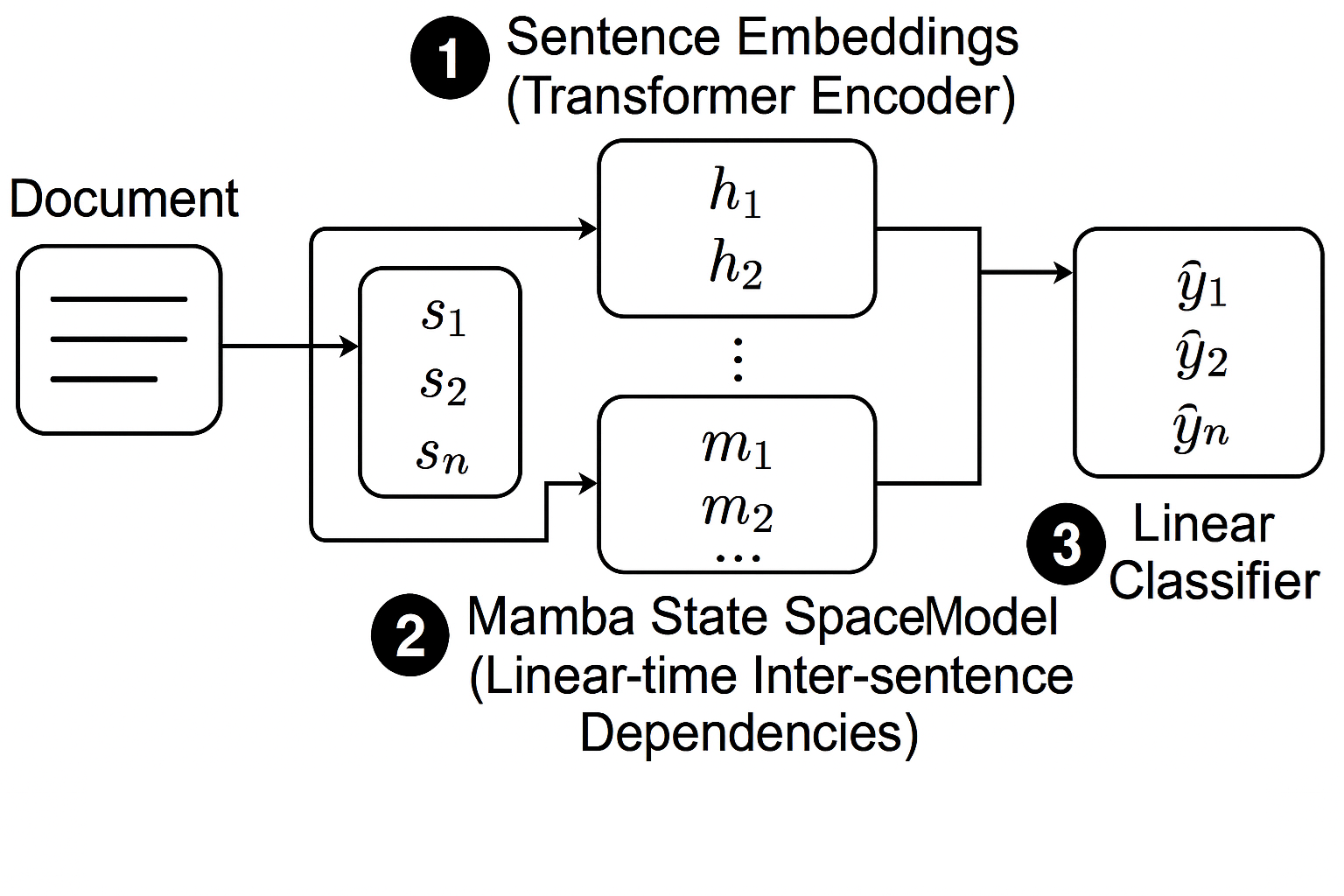}
    \caption{Overview of the proposed Mamba–Transformer hybrid architecture for extractive summarization.}
    \label{fig:architecture}
\end{figure}
\subsection{Problem Formulation}
We formulate extractive summarization as a document-level sentence classification task.  
Given a document $D = \{ s_1, s_2, \dots, s_n \}$ consisting of $n$ sentences, our objective is to classify each sentence as relevant ($y_i = 1$) or not relevant ($y_i = 0$) for the summary, while considering inter-sentence relationships within the document context.

\subsection{Architecture Overview}
Our approach consists of three key components that process documents at the sentence-sequence level.

\subsubsection{Sentence Encoding with Transformers}
For each document $D = \{ s_1, s_2, \dots, s_n \}$, we first apply a pre-trained transformer (BERT) to generate semantic embeddings $[{\tt CLS}]$ for each sentence independently:
\[
h_i = \mathrm{Transformer}(s_i) \in \mathbb{R}^d
\]
where $h_i$ represents the sentence-level semantic encoding for sentence $s_i$.

\subsubsection{Document-Level Sequential Processing with Mamba}
We then construct a sequence of sentence embeddings  
$H = [h_1, h_2, \dots, h_n] \in \mathbb{R}^{n \times d}$  
and process this sequence through the Mamba state space model to capture inter-sentence dependencies within the document:
\[
M = \mathrm{Mamba}(H) = [m_1, m_2, \dots, m_n] \in \mathbb{R}^{n \times d}
\]
This step enables the model to capture document structure and sentence relationships while maintaining linear computational complexity $\mathcal{O}(n)$ with respect to the number of sentences.

\subsubsection{Binary Classification for Sentence Relevance}
Finally, we apply a linear classifier to each position in the Mamba output sequence to predict sentence-level relevance:
\[
\hat{y}_i = \sigma (W \cdot m_i + b)
\]
where $m_i$ is the Mamba encoding for sentence $i$, $W \in \mathbb{R}^d$ is the weight matrix, $b$ is the bias term, and $\sigma$ is the sigmoid function.  
This produces relevance probabilities for each sentence in the document.

\section{Experimental Setup}
\subsection{Dataset}
 Due to the GPU limitations, we evaluated our Mamba-Transformer model on three diverse datasets using 200 documents each, representative of realistic low-resource scenarios where large-scale annotation and computational resources are constrained.
\textbf{Datasets}:
\begin{itemize}
    \item CNN/DailyMail \cite{hkgeksb15}: News articles with multi-sentence highlights
    \item DebateSum \cite{rb20}: Debate evidence requiring argumentative structure understanding
    \item ArXiv \cite{cdkbkcg18}: Scientific papers with abstracts, ideal for evaluating scalability
\end{itemize}
\textbf{Preprocessing:} We generated sentence-level binary labels using a greedy ROUGE-2 maximization approach. For dataset diversity, we applied K-means clustering (500 clusters) using TF-IDF representations, then randomly selected 200 documents. Text was segmented using NLTK, tokenized with HuggingFace, and truncated to 128 tokens. Documents were split 120/40/40 (train/val/test) at document level to prevent data leakage.

\begin{table}[t]
  \centering
  \resizebox{.95\columnwidth}{!}{%
  \begin{tabular}{lccc}
    \hline
    \textbf{Dataset} & \textbf{Avg doc length (tokens)} & \textbf{Avg sents/doc} & \textbf{Positive (\%)} \\
    \hline
    CNN/DailyMail & 781   & 29.4 & 18.2 \\
    DebateSum     & 643   & 24.1 & 22.7 \\
    ArXiv         & 1{,}247 & 41.8 & 16.9 \\
    \hline
  \end{tabular}%
  }
  \caption{Dataset statistics.}
  \label{tab:dataset-stats}
\end{table}

\subsection{Implementation Details}
Our architecture consists of a BERT-base-uncased encoder (110M parameters, 768-dimensional embeddings) followed by a Mamba-130M state space model (NF4 quantized, $d_{\mathrm{state}}=16$, $d_{\mathrm{conv}}=4$) and a linear classifier mapping from 768 to 1 with a sigmoid activation. Training was performed using the Adam optimizer (learning rate $1\times 10^{-5}$), binary cross-entropy loss, batch size of 1 with gradient accumulation, for 2 epochs. We applied  0.2 dropout and gradient clipping with $\mathrm{max\_norm} = 1.0$. Data preprocessing used the BERT tokenizer with a maximum length of 128 tokens per sentence, ROUGE-2 greedy labeling for supervision, and no document truncation. Experiments were conducted on an NVIDIA A100 GPU (40GB) using PyTorch 2.6.0+cu124 and HuggingFace Transformers 4.35.0, with random seeds $\{42, 123, 456\}$. Evaluation was performed using ROUGE-1, ROUGE-2, and ROUGE-L F1 scores, with paired t-test significance testing.

\section{Results}
\subsection{Overall Performance}
Across three domains, MAMBA-BERT consistently outperformed transformer-only baselines in ROUGE scores. On CNN/DailyMail (news), it achieved 0.61/0.51/0.53 (ROUGE-1/2/L), a significant gain over BERTSUM (+0.17 ROUGE-1) and MATCHSUM (+0.21 ROUGE-1). On DebateSum (argumentative), improvements were smaller but consistent. The largest gains occurred on ArXiv (scientific), where MAMBA-BERT surpassed BERTSUM by +0.23 ROUGE-1 and MATCHSUM by +0.56. Statistical significance testing confirmed these improvements are not due to random variation, with MAMBA-BERT significantly outperforming BERTSUM on ROUGE-1 across all domains: CNN/DailyMail (news) ($t = 5.503$, $p < 0.001$, Cohen's $d = 1.395$, $n = 40$), DebateSum ($t = 9.6$, $p < 0.001$, Cohen's $d = 2.702$, $n = 40$), and ArXiv ($t = 28.09$, $p < 0.001$, Cohen's $d = 6.183$, $n = 40$). These results indicated both statistical significance and large practical effect sizes across datasets. We attempted to evaluate MemSum as a recent extractive baseline; however, under our low-resource setting (200 documents), the model produced unstable and illogical outputs. We therefore excluded MemSum from quantitative comparison, noting this limitation

\subsection{Efficiency and Long-Document Advantage}
On news summarization, MAMBA–BERT reduced inference time by 24–27\% compared to baselines while maintaining linear complexity scaling. On ArXiv, which contains the longest documents, our model not only processed full inputs without truncation but also delivered the highest relative quality gains, confirming its suitability for long-document summarization.
\subsection{Low-Resource Robustness}
All experiments were conducted with only 200 documents per dataset. Despite this constraint, MAMBA–BERT maintained strong performance across domains, demonstrating effectiveness in low-resource conditions.
\begin{table}[t]
  \centering
  \resizebox{.95\columnwidth}{!}{%
  \begin{tabular}{lcccc}
    \hline
    \textbf{Model} & \textbf{ROUGE-1} & \textbf{ROUGE-2}  & \textbf{ROUGE-L} & \textbf{Time (s/sample)} \\
    \hline
    \multicolumn{5}{l}{\textbf{CNN/DailyMail}} \\
    \hline
    MAMBA-BERT & \textbf{0.61} & \textbf{0.51} & \textbf{0.53} &  \textbf{13.7} \\
    BERTSUM     & 0.44 & 0.28 & 0.33 & 18.0 \\
    MATCHSUM    & 0.40 & 0.18 & 0.27 &  18.6 \\
    \hline
    \multicolumn{5}{l}{\textbf{DebateSum}} \\
    \hline
    MAMBA-BERT & \textbf{0.57} & \textbf{0.54} & \textbf{0.56}& 0.65 \\
    BERTSUM     & 0.50 & 0.35 & 0.40 & 0.52 \\
    MATCHSUM    & 0.46 & 0.34 & 0.37 &  0.12 \\
    \hline
    \multicolumn{5}{l}{\textbf{ArXiv}} \\
    \hline
    MAMBA-BERT & \textbf{0.73} & \textbf{0.64} & \textbf{0.64} & 8.14 \\
    BERTSUM     & 0.50 & 0.26 & 0.27 & 3.67 \\
    MATCHSUM    & 0.17 & 0.04 & 0.12 & 0.45 \\
    \hline
  \end{tabular}}%
  \caption{ROUGE scores, and inference times.}
  \label{tab:results}
  \vspace{0.1cm}
  {\footnotesize 
  Statistical significance testing ($p < 0.001$, paired t-test, $n = 40$): MAMBA-BERT vs BERTSUM on ROUGE-1 across all datasets.}
\end{table}

\section{Error Analysis}
Manual examination of the lowest-performing cases revealed two primary failure modes:
\begin{itemize}
\item \textbf{Over-selection of irrelevant details (100\%):} The model consistently selected less important background information or excessive details instead of core content. 
\item \textbf{Missing critical information (75\%):} - Named entity omissions: Missing names, organizations, digital platforms - Missing main events: Key actions, decisions, or announcements - Missing contextual information: Explanatory context and characterizing quotes - Missing attribution details: Source information and age details.
\end{itemize}
The error analysis showed a clear trade-off: The model was good at identifying semantically relevant content but struggled with importance ranking and entity prioritization. This suggests that MAMBA's sequential modeling might need enhancement with importance-aware attention or entity-aware mechanisms.

\section{Conclusion, Limitations, and Future Work}
We introduce the first Mamba-Transformer hybrid for extractive summarization, combining transformer-based semantic encoding with the linear-time processing of Mamba state space models. This approach processes full-length documents without truncation and achieves strong results in low-resource settings, outperforming BERTSUM and MATCHSUM across three domains (+0.23 ROUGE-1 on ArXiv; $p<0.001$ on news) while reducing inference time by up to 27\%. Paired $t$-tests confirm that these gains are statistically significant rather than random. Our findings highlight both cross-domain generalization and efficiency, demonstrating the broad applicability of the method.

Nonetheless, several limitations remain. Importance ranking and entity prioritization are not explicitly modeled, leaving room for importance-aware or entity-aware enhancements. Our experiments were limited to 200 documents per dataset due to resource constraints; larger-scale studies would strengthen validation. In addition, comparisons focused on earlier baselines selected for constrained settings, and preliminary tests with newer extractive methods did not yield competitive results in our low-resource scenario. Expanding baseline coverage, scaling training data, and exploring hybridization with abstractive techniques represent promising next steps.

We hope this opens new directions for hybrid architectures in efficient neural summarization in domains such as scientific literature, newsrooms, and policy debates, where processing long documents under limited resources is essential.

\bibliography{bibtex}

\end{document}